\documentclass[letterpaper, 10 pt, conference]{ieeeconf}
\IEEEoverridecommandlockouts

\usepackage{booktabs}
\usepackage{amsmath,amssymb,amsfonts}
\usepackage{algorithmic}
\usepackage{graphicx}
\usepackage{textcomp}
\usepackage{tabularx}
\usepackage{xcolor}
\usepackage{hyperref}
\usepackage{ragged2e}
\justifying

\title{\LARGE \bf
See What I Mean?  Expressiveness and Clarity in Robot Display Design}
\author{
Matthew Ebisu$^{1, \dagger}$, Hang Yu$^{2, \dagger}$, Reuben Aronson$^{3}$, Elaine Short$^{4}$%
\thanks{$\dagger$ These authors contributed equally to this work.}
\thanks{$^{1}$M. Ebisu is a graduate student researcher with the AABL Lab, Tufts University, Medford, MA, USA. {\tt\small matthew.ebisu@tufts.edu}}%
\thanks{$^{2}$H. Yu is a PhD student with the AABL Lab, Tufts University, Medford, MA, USA. {\tt\small hang.yu625917@tufts.edu}}%
\thanks{$^{3}$R. Aronson is a postdoctoral researcher with the AABL Lab, Tufts University, Medford, MA, USA. {\tt\small reuben.aronson@tufts.edu}}%
\thanks{$^{4}$E. Short is a faculty member with the AABL Lab, Tufts University, Medford, MA, USA. {\tt\small elaine.short@tufts.edu}}%
}

\usepackage{xstring}

\DeclareUnicodeCharacter{2009}{\,}  

\begin{document}

\maketitle

\begin{abstract}


Non-verbal visual symbols and displays play an important role in communication when humans and robots work collaboratively. 
However, few studies have investigated how different types of non-verbal cues affect objective task performance, especially in a dynamic environment that requires real-time decision-making. In this work, we designed a collaborative navigation task where the user and the robot only had partial information about the map on each end and thus the users were forced to communicate with a robot to complete the task. We conducted our study in a public space and recruited 37 participants who randomly passed by our setup. Each participant collaborated with a robot utilizing either animated anthropomorphic eyes and animated icons, or static anthropomorphic eyes and static icons. 
We found that participants that interacted with a robot with animated displays reported the greatest level of trust and satisfaction; that participants  interpreted static icons the best; and that participants with a robot with static eyes had the highest completion success.
These results suggest that while animation can foster trust with robots, human-robot communication can be optimized by the addition of familiar static icons that may be easier for users to interpret.
We published our code, designed symbols, and collected results online at: \url{https://github.com/mattufts/huamn\_Cozmo\_interaction}. 
\end{abstract}


\section{Introduction}

Effective communication is critical in human-robot interaction (HRI), particularly in collaborative tasks where humans and robots must work together to achieve shared goals. For robots to communicate their intentions effectively, their actions and signals must be readable, predictable, and intuitive to humans \cite{s21175722}. In this context, visual displays play a vital role as they serve as the primary interface for robots to convey their internal states, intentions, and guidance in real-time \cite{10.1145/3317325}\cite{fang2025demonstration}. 
When robots express intent clearly, they reduce the cognitive load on human collaborators, enabling faster and more accurate decision-making during tasks \cite{10.3389/frobt.2022.754955}.
\begin{figure}
    \centering
    \includegraphics[width=0.95\linewidth]{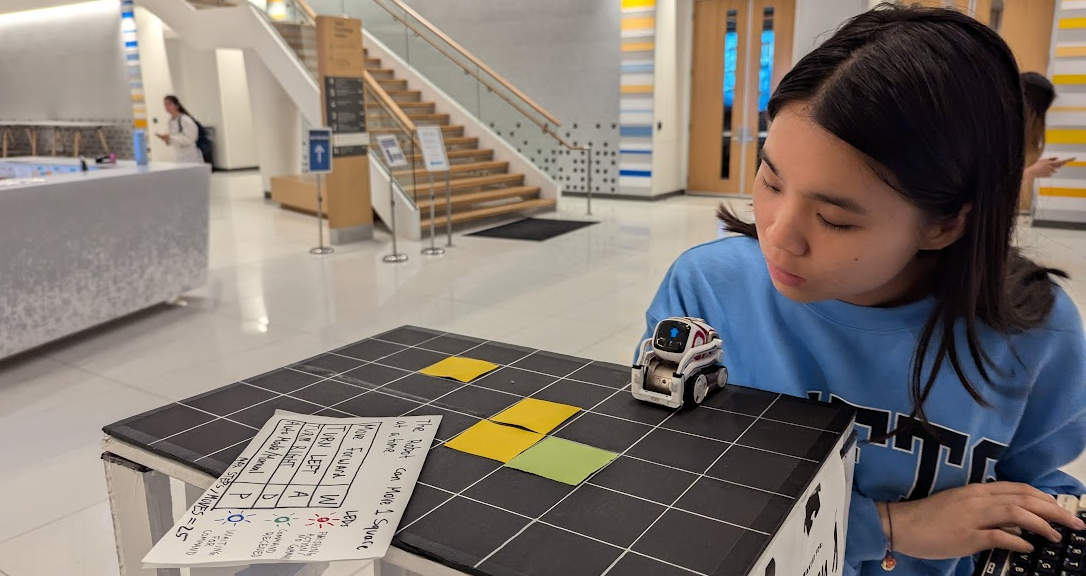}
    \caption{The study was set in a university atrium on the first floor. 
    Participants were recruited from people who passed by our setup.
    The task was a collaborative navigation task, where the obstacles were invisible to participants and the hazards were invisible to the robot.} 
    \label{fig:enter-label}
\end{figure}

\indent Animation has emerged as a powerful tool in robotics for enhancing communication \cite{10.1145/3317325}. Animated displays add dynamism and temporal cues to robot communication, making their actions and signals more expressive and engaging. Animation techniques, such as blending, timing, and motion fluidity \cite{10.7230/KOSCAS.2017.49.677}, can create a sense of purposeful movement that humans naturally interpret as indicative of internal states or goals. As a result, animation improves the readability and legibility of a robot’s actions, particularly in scenarios requiring quick and intuitive human responses
\cite{10.1145/2157689.2157814}.  \\  \indent Anthropomorphism in animation -- the application of human-like characteristics to robots such as expressive eyes or gaze direction-- is another widely explored means of enhancing communication in HRI \cite{10.1145/3543174.3546841}. Anthropomorphic features tend to evoke natural human responses, foster trust, relatability and perceived competence, and improve the user's ability to interpret robot actions by mimicking familiar social cues such as attention, emotion, and intent \cite{10.1145/3319502.3374839}.

Although prior research has explored animation and anthropomorphism in robotics, much of the prior work evaluates these elements in isolation. Few studies have examined how these characteristics work in tandem or compared them systematically to non-animated (static) and non-anthropomorphic (icons) displays. Addressing this gap is crucial for understanding whether animation and anthropomorphism, when combined, offer synergistic benefits in HRI. In particular, a systematic comparison of animated anthropomorphic displays (e.g., expressive eyes) against static and symbolic alternatives (e.g., icons) could provide insights into how these features impact task performance and human-robot communication.

In this work, we compared icons and anthropomorphic displays across static and animated conditions in a collaborative task.
The task was a navigation task where participants guided a robot through a maze. The obstacles in the maze were invisible to participants, and the fire hazards were invisible to the robot.  
 We 
 recruited 37 participants in an atrium of a university building. 
Our analysis incorporates both objective (task completion and display interpretation) and subjective metrics (self-reported trust, satisfaction, and understandability). 
We found that the participants who worked with the robot utilizing animated display types reported a higher level of trust and satisfaction than with the robots employing static displays. 
The participants who worked with robots utilizing static eyes, however, achieved the highest success rate in performing the collaborative task. 
Moreover, we found a low correlation between participants' self-reported understandability and their actual performance in interpreting the expressions. 
All expressions we designed, codes we used, and data we collected are online at \href{https://github.com/mattufts/huamn\_Cozmo\_interaction}{our GitHub repository}. 

Our contribution is a robot-in-the-wild study that objectively and subjectively compares four non-verbal modalities with a collaborative task.
We found inconsistencies between users' subjective feelings and their actual performance, highlighting the importance of including both subjective and objective measurements.  
Our results provide a guideline for applying different visual cues to different scenarios.
when testing or developing new communication modalities.



\section{Background}

Non-verbal expressions have been widely used in human-robot communication 
\cite{10.1145/1514095.1514110}\cite{1545011}\cite{10.1145/1514095.1514109}\cite{8525767}, and have been shown to be successful in many applications, such as successfully influencing task performance and perception of robots \cite{10.1145/1514095.1514110}\cite{yu2024much}. 
Human participants improved their task performance when robots used non-verbal cues \cite{s21175722}, even when they did not explicitly recognize or report the cues \cite{10.1145/1514095.1514110}.
Our study is centered on the following designs in robotics: Anthropomorphism \cite{10.3389/frobt.2022.848295}\cite{10.1145/3171221.3171286}\cite{ROESLER2024108008}, Animation \cite{10.1145/2157689.2157814}\cite{10.1145/3317325} and Non-verbal Communication in Collaborative Tasks \cite{1545011}\cite{10.1145/1514095.1514109}\cite{10.1145/1514095.1514110}.

\subsection{Anthropomorphism in Robotics}
Anthropomorphism in robotics refers to the design practice of giving robots human-like characteristics, such as facial features or gestures, to make them more relatable, intuitive and emotionally engaging \cite{10.5555/515422}\cite{saldien2014motion}. 
This includes ensuring that robot outward expressions, such as facial movements \cite{8525767}, gaze shifts \cite{10.1145/3543174.3546841}, and gestures \cite{9900752}, align with how humans naturally interpret social signals \cite{1545011}\cite{10.5555/515422}\cite{yu2021active}\cite{tan2020top}.
Humans naturally attribute human characteristics to non-human objects, including robots, which can make them more appealing and engaging \cite{8525767}. 
 Anthropomorphic features in product design can attract people, as even minimal social cues are enough to trigger anthropomorphism \cite{10.3389/frobt.2023.1178433}. 
The design of the eyes and face of a robot significantly influences how people perceive and interact with it. Robots with more lifelike eyes are perceived as more personable and more suitable for the home environment \cite{10.3389/frobt.2023.1178433}.  
The results of one notable study showed that robots with lifelike eyes were rated higher on personable traits (e.g., sensible, loyal, thoughtful) and were deemed more suitable for home environments, while abstract or no eyes were associated with professional traits but were perceived as cold and harder to interact with \cite{10.3389/frobt.2022.848295}. 
Not all studies, however, are in agreement that the addition of eyes improves human-robot communication and collaboration \cite{PILACINSKI2023e18164}.

\subsection{Animation in Robotics} 

Animation is an important advancement in robotic design, enhancing user engagement and social relatability \cite{Ribeiro2020ThePO}.  Researchers use animation to try to make robot behaviors more readily apparent to human users \cite{6281390} \cite{1545011}.
The integration of motion-based expressive elements enables robots to convey intent dynamically, often eliciting stronger human responses compared to static imagery. Past studies has described robotic animation as a mechanism for simulating life, wherein robots convey internal states, motivations, and responsiveness to maintain interactive autonomy \cite{10.1145/2157689.2157814}. Some researchers have concluded that animation enhances the perception of movement and intent, making emotionally expressive eye gestures easier to process quickly (The 12 Principles of Animation) \cite{9900752}.
The introduction of continuous, dynamic signals could increase cognitive processing demands, particularly in time-sensitive decision-making scenarios where users must extract clear, actionable information \cite{10.1145/3317325}.
From an operational standpoint, it is critical to assess whether the illusion of life, as conveyed through animation, enhances or impairs the robot’s ability to provide precise navigational cues.  

\subsection{Non-Verbal Communication in Collaborative Tasks}

Previous work in this field has highlighted the importance of readability and social understanding in the design of social robots \cite{1545011}\cite{10.5555/515422}\cite{fang2025charmconsideringhumanattributes}\cite{yu2023thumbs}. 
To foster seamless interaction, robots must achieve real-time responsiveness, coherence in their movements, synchrony between actions and expressions, and expressive versatility to reflect emotions \cite{Breemen2004BringingRT}. By making these social cues readable, robots can support humans' social ability to understand others, thereby improving the quality and naturalness of interactions \cite{5650128}.
Building on the importance of anthropomorphism in fostering social interaction, studies examined how non-verbal cues, particularly gaze, could help humans accurately identify a robot’s intentions \cite{10.1145/3543174.3546841}. 
Semioticians have classified signs into three categories \cite{crow2010visible}:
Symbol, which is an arbitrary signifier that is learned such as road signs; Icon, which is a signifier that resembles or mimics the signified such as a cartoon; and Index, which is a signifier that is not arbitrary such as a footprint. They can include a wide range of human-robot interaction cues, such as directional arrows or familiar traffic signals like red and green lights. Semiotic signs and symbols are useful as a framework to model human-robot interactions \cite{10.1007/978-3-031-38454-7_10}.

Our work differs from prior work in that we conducted our study in a non-lab setup with a collaborative task, while collecting results using both qualitative and quantitative measures. 
Unlike prior work which often examined animation and anthropomorphism in isolation, our study directly tests the interrelated functional impact of animation, anthropomorphism and icons on task performance and human perception using both objective and subjective metrics. 

\section{Methodology}
We conducted a public space study to determine how display modalities could affect human robot communication in a shared navigation task.
Participants were volunteers who passed by our study setup.
In this task, both humans and the robot only had partial information, requiring effective communication for successful maze completion.
The public space setting ensured a valid environment for evaluating how non-verbal robotic displays affect communication in real-time and in the real world.  

\subsection{Study Design}
Our goal was to investigate how different visual displays influence human-robot communication in a collaborative maze navigation task.
We used a Cozmo robot with four modalities: animated eyes, animated icons, static eyes, and static icons. 
We conducted the study in a university atrium.
We recruited participants by asking people who passed by our study setup. 
We randomly assigned them to one of two groups: Animated Displays or Static displays. 
Cozmo and the participants could each see part of the map environment and were thus forced to communicate to complete the task. 
We collected both objective results (task performance, display interpretation) and subjective results (self-reported understandability, trust in robotics, and satisfaction with the robot). 
The study was approved by the university Institutional Review Board (IRB).
We further introduce the details of the display modalities, the task, and the study procedure in the following sections.  





\subsection{Display Modalities}

We selected four distinct visual display categories: animated eyes, animated icons, static eyes, and static icons. 
Within the four categories, we created eight communicative displays for each category that helped guide and inform each participant about the invisible maze environment.  These were: an \textit{Idle State Display} for when the robot was awaiting commands; an \textit{Affirmation Display }to indicate that a user's command matched the robot's intent; three \textit{Directional Displays} (Left, Right, Forward) for navigation cues; a \textit{Wall Impact Display} for collisions; a \textit{Hazard Impact Display} for environmental dangers; and a \textit{Goal Success Display} to indicate task completion.
One example of our designed expressions is shown in Figure 2.
These designs served as the robot’s primary communication and were designed to convey directional intent, environmental reactions, and task alignment cues.
Each visual category was systematically structured across anthropomorphic and icon display types, ensuring consistency in meaning and motion across all conditions.

\begin{figure}
\label{fig:Display_Images}
    \centering
    \includegraphics[width=0.95\linewidth]{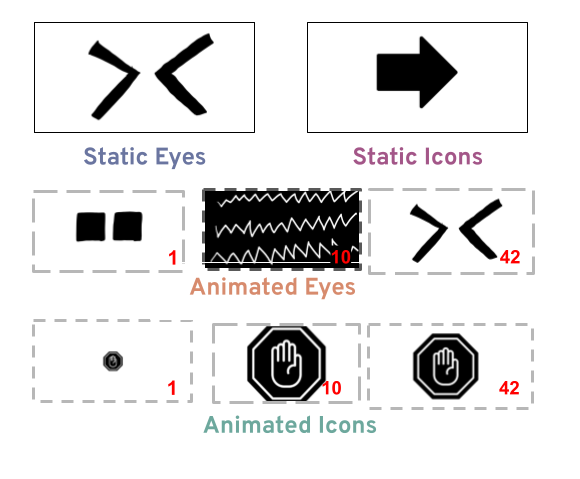}
    \caption{“Wall Impact Display” for collisions. Static Eyes and Static Icons displayed the shown image for the entire duration. Animated Eyes and Animated Icons depict images taken from the animation display sequence at frames 1, 10, and 42. 
    Additional details and examples are at our \href{https://github.com/mattufts/huamn\_Cozmo\_interaction}{GitHub}
    }
\end{figure}

 The icons were designed to resemble street signs and emoticons, using simple monochrome graphics, adapted from The Noun Project (https://thenounproject.com/) to ensure familiarity. The eyes, inspired by Cozmo’s original expressive designs, utilized simplified gaze-based expressions to communicate intent. Both animated and static displays maintained uniformity in their representation, with equivalent keyframe counts and motion behaviors across all conditions.

\subsection{Collaborative Navigation Task}
\begin{figure}
    \centering
    \includegraphics[width=0.95\linewidth]{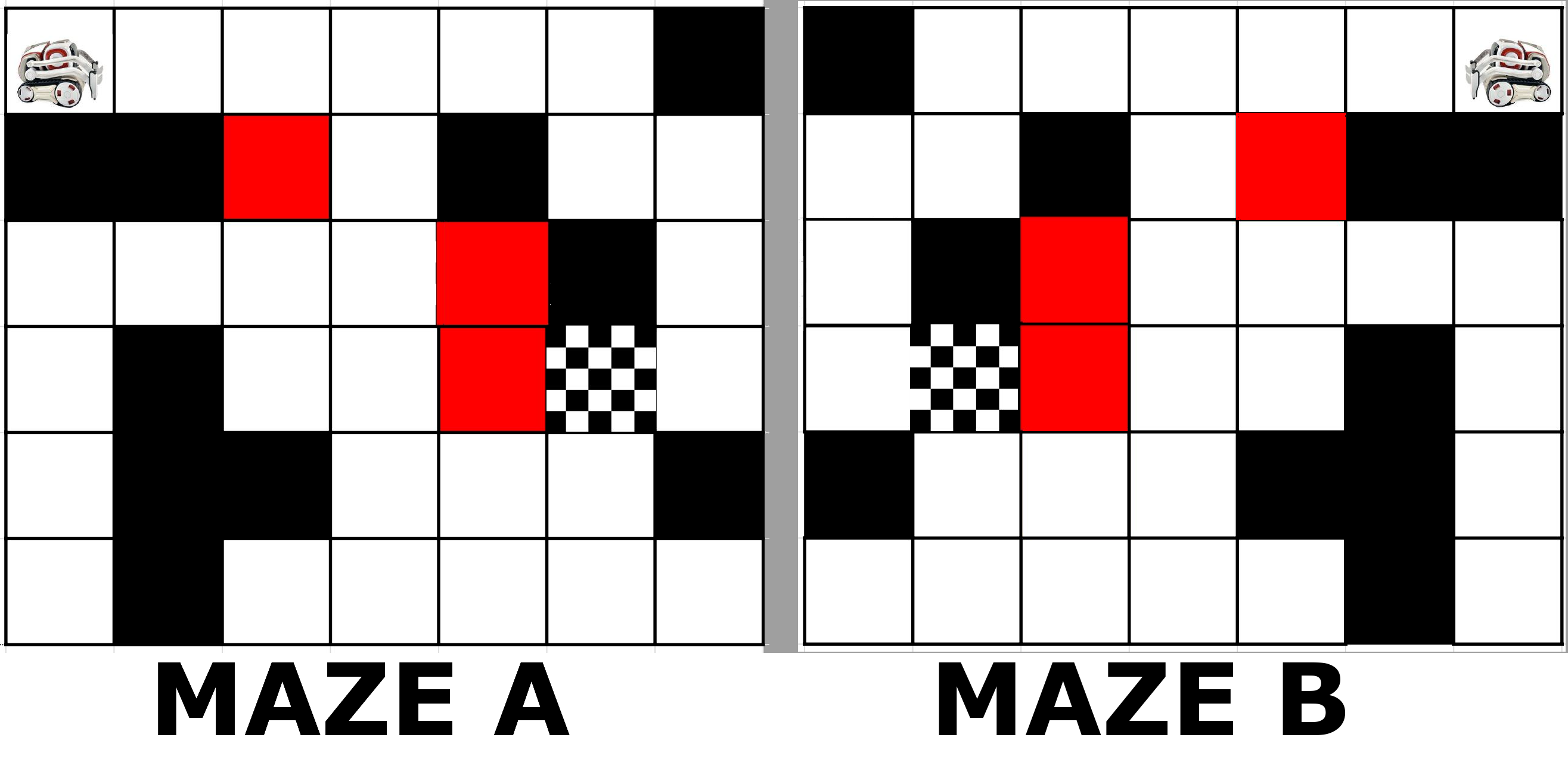}
    \caption{
    Maze layout for the collaborative task.  Black squares were walls visible only to the robot, and red squares were environmental hazards visible only to the participants. The checkered flag was the goal and was visible to both the participants and the robot. Each participant used both mazes in a random order.}
    \label{fig:MazeAB}
\end{figure}

The task we designed was intended to be a uniquely collaborative navigation task between a human and a robot, mirroring on an experimental level a real-world collaborative rescue mission. Our participants were instructed at the outset that they were to work with the robot as a team in their mission.
The goal was to navigate the Cozmo robot to move from the initial position to the goal position. 
We used two mazes in our study, shown in  \autoref{fig:MazeAB}. 
The black grids are obstacles and the red grids are fire hazards. 
The obstacles were invisible to the participants, and the fire hazards were invisible to the Cozmo. 
The participants ran trials on both Maze A and Maze B in a random order. 
The two mazes were mirrors of each other to ensure that all participants encountered a standardized level of difficulty regardless of assignment.

Participants were considered to have successfully completed the maze if they reached the exit within the maximum allowed steps (25 moves) and a total health point value that was above 0.
The initial total heath point value for Cozmo was set to 100.
Hitting a wall would result in losing 10 health points from their total, and hitting a fire hazard would result in losing 20 heath points.
All visual displays were displayed on Cozmo's screen. 
We had a voice interface implemented for controlling the robot but we decided to use a keyboard interface in the end for reliability in the public space. 

The maze was placed at eye level on a desk, allowing participants to clearly observe Cozmo’s OLED display while tracking the robot’s movements.  Participants were instructed that they could walk around the perimeter of the maze, ensuring full visibility of Cozmo’s real-time feedback.
Each participant used both mazes in their experimental trial.

\subsection{Procedure}
We set up our study in an atrium of a university building. 
We recruited participants who were attracted by our setup and expressed interest in participating. 
After filling out a consent form, each participant received a brief orientation outlining the robot’s capabilities and the interaction protocol. 
Participants were tasked with guiding the Cozmo robot through two separate maze-like environments designed to simulate building escape scenarios. 
They were informed that the maze layout itself would remain unseen, but hazardous zones (yellow) and goal areas (green) would be visible. 
Participants could issue movement commands through a keyboard, directing Cozmo to move.
Participants were required to continuously monitor Cozmo’s screen to make decisions throughout the task.  
Cozmo provided visual feedback to assist in navigation.
After each movement, Cozmo displayed a visual cue on its face corresponding to the suggested next action. 
As the protocol was that only the human would have visual information about the hazards, the next action suggested by Cozmo was not guaranteed to be correct.
If they failed to input a command after four seconds, Cozmo autonomously executed a movement, continuing along its predetermined navigation path. 
The task automatically terminated if the participant exceeded 25 steps. 

Following the completion of both maze trials, participants filled out a post-task questionnaire.
The questionnaire employed an 11-point Likert scale to measure trust \cite{bagheri2022transparent}, satisfaction \cite{bagheri2022transparent}, and understandability \cite{inbook}.
After completing the post-trial questionnaire, participants were asked to complete a second task we informally called the “Interpretation Game.” It consisted of a list of display signals corresponding to their assigned group of either Animated Displays or Static Displays. They were asked to interpret each signal and provide their best understanding of what the robot had tried to communicate via each display during the study. The evaluation of the  Interpretation Game was conducted by the author and two other assistants.    Each response was evaluated on an accuracy scale ranging from 0 to 2, where 0 indicated an incorrect interpretation, 1 represented a neutral or ambiguous response, and 2 signified a completely accurate interpretation. Participants who did not recall seeing a particular signal during the study were instructed to write, “I didn’t see the icon." which did not factor into our calculation.  Finally, the different graders’ scores for each participant’s responses were averaged for each of the responses to each display.

\begin{figure}
    \centering
    \includegraphics[width=0.95\linewidth]{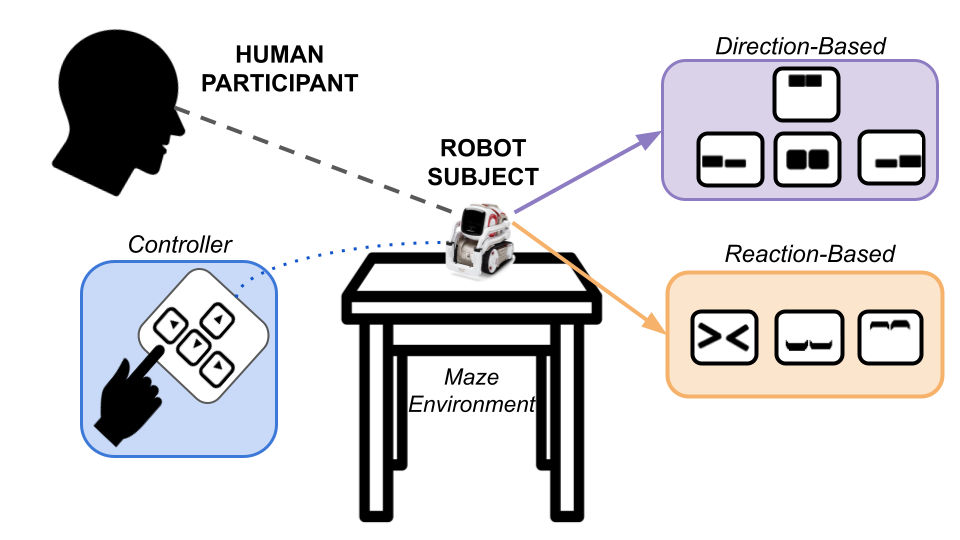}
    \caption{The Experimental setup involving a human participant, Cozmo robot, and directional/reaction-based cues.}
    \label{fig:maze}
\end{figure}


\section{Results}
In this work, we aimed to explore how different visual modalities would influence human-robot communication in a collaborative task. We set up a maze navigation experiment in a public setting with 37 non-expert volunteers. The robot was programmed with four distinct visual display modalities: animated eyes, animated icons, static eyes, and static icons. The volunteers were randomly assigned to one of two groups: Animated Displays or Static Displays. The human and the robot had to communicate with each other in order to complete the task. 


\subsection{Participants with static eyes modality had the highest percentage of task successes to task failures}

\begin{figure}
    \centering
    \includegraphics[width=0.95\linewidth]{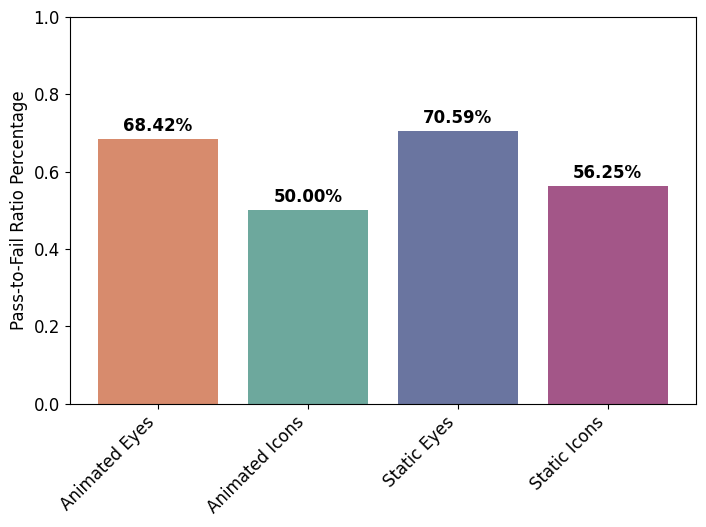}
    \caption{ Percentages of trials completed successfully for maze completion across modalities: Static Eyes (70.59\%), Animated Eyes (68.42\%), Static Icons (56.25\%), and Animated Icons (50.00\%). Ratios reflect the percentage of successful completions out of total trials for each display condition.}
    \label{fig:suc}
\end{figure}

We evaluated the performance of human participants in the maze navigation task with Cozmo across the four visual communication modalities: Static Eyes, Animated Eyes, Static Icons, and Animated Icons.  Performance was measured using the pass-to-fail ratio—the percentage of trials in which participants successfully reached the end of the maze out of the total number of attempts. 
The pass-to-fail ratios for each modality are presented in \autoref{fig:suc}. Participants using the Static Eyes modality achieved the highest pass-to-fail ratio, successfully completing the maze in 70.59\% of trials. The Animated Eyes modality followed with a pass-to-fail ratio of 68.42\%. In comparison, the Static Icons modality resulted in a pass-to-fail ratio of 56.25\%, while the Animated Icons modality had the lowest pass-to-fail ratio at 50.00\%.

\subsection{Participants’ subjective understandability did not necessarily indicate their objective understandablity}

\begin{figure}
    \centering
    \includegraphics[width=0.95\linewidth]{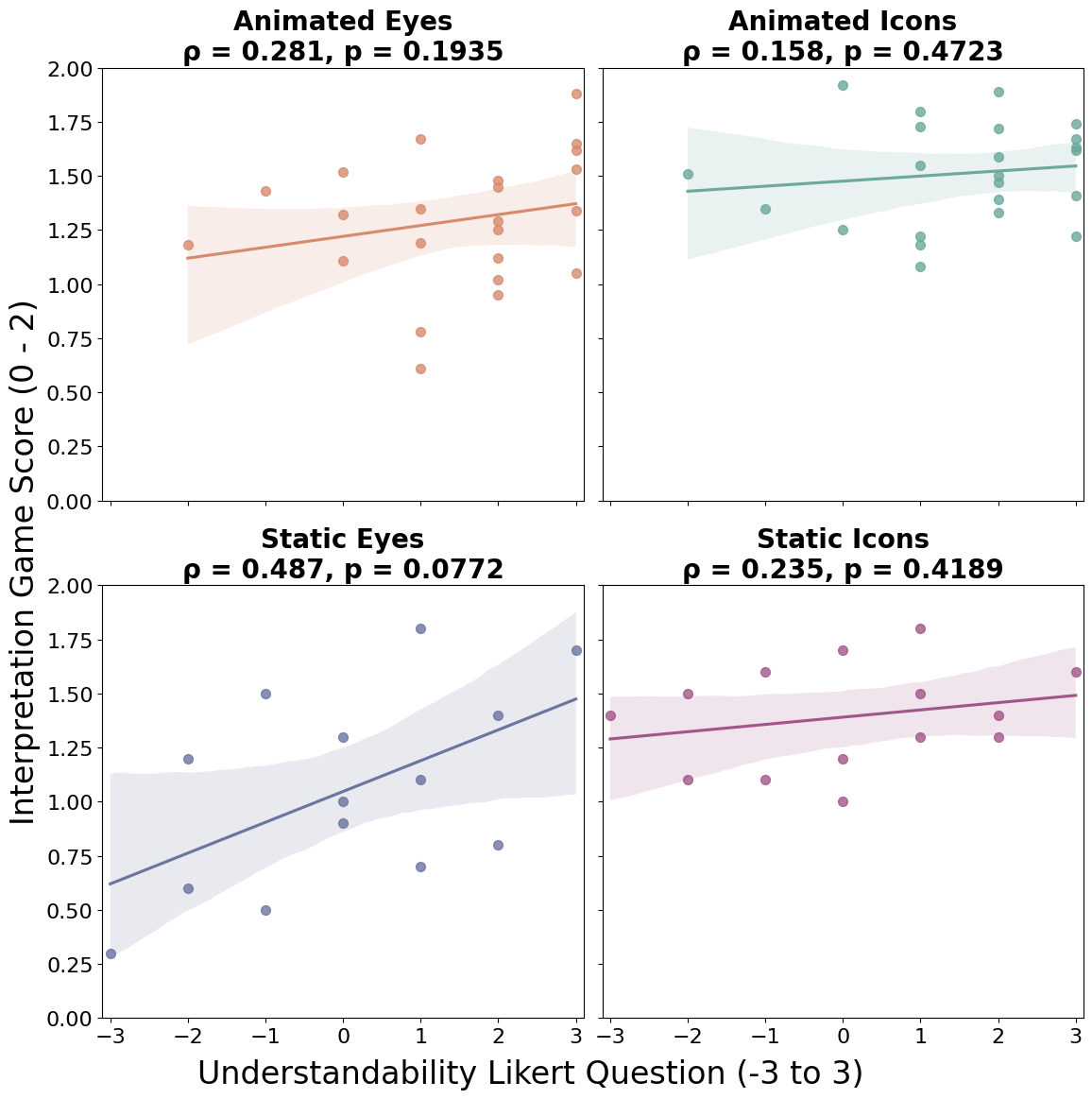}
    \caption{ Interpretation Game score and self-reported understandability for each participant. The y-axis represents the Interpretation Game score and the x-axis is the self-reported understandability score. 
    We calculated a Spearman correlation for each category, shown on the top of each figure. 
    No statistical significance was found across the Interpretation Game to self-reported understandability.}
    \label{fig:und}
\end{figure}

To assess the relationship between participants’ subjective understanding and their objective performance, we compared self-reported understandability scores with their performance in the Interpretation Game. Each participant completed Likert-style self-assessments in response to the post-survey statement
on a scale from -3 (strongly disagree) to +3 (strongly agree).
Scores from the Interpretation Game were calculated with the average of each participant’s score per display condition, with possible scores ranging from 0 to 2 per trial. 
We show their self-reported understandability and their Interpretation Game performance in \autoref{fig:und}.
Each subplot displays the relationship between participants’ self-reported understandability (x-axis) and their average Interpretation Game performance (y-axis) across one of the four display modalities. The data points represent individual participants. The solid regression line indicates the general trend in the data, estimated using linear regression. The shaded region surrounding the line represents the 95\% confidence interval of the linear fit.

We calculated the mean scores per modality for their Interpretation Game: Animated Eyes 0.888, Animated Icons 1.321, Static Eyes 0.839 and Static Icons at 1.403. 
We found that static icons are most interpretable, using a one-way ANOVA which confirmed a significant difference across modalities \((F(3, N-4) = 9.06, p < .001)\).
To evaluate whether this objective performance was reflected in participants’ self-perceived understanding, we computed Spearman rank correlations between the average Interpretation Game scores and the raw (-3 to 3) understandability ratings for each condition. Across all four modalities, we found no statistically significant correlation between participants’ self-assessments and their objective interpretation scores: Animated Eyes ($r_s$ = 0.281, p=0.1935), Animated Icons ($r_s$ =0.158, p = 0.4723), Static Eyes ($r_s$ = 0.487, p = 0.0772) and Static Icons ($r_s$ = 0.235, p = 0.4189) 


\subsection{Participants rated the animated displays as more trustworthy and rated higher scores for satisfaction}

\begin{figure}
    \centering
    \includegraphics[width=0.95\linewidth]{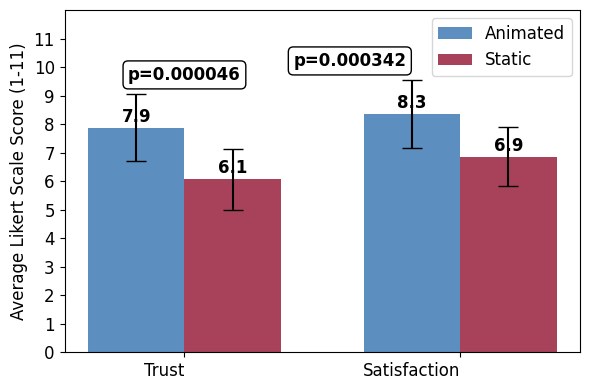}
    \caption{
    Average Trust and Satisfaction scores by modality. Animated displays rated significantly higher than Static displays (Trust: $M=7.9$ vs.\ $6.1$, $p<0.001$; Satisfaction: $M=8.3$ vs.\ $6.9$, $p<0.001$). Error bars represent standard deviation.
}
    \label{fig:trust & stat}
\end{figure}

While we showed that static expressions facilitate participants to achieve better performance on a collaborative task, including animated expressions is still beneficial, especially in human-robot communication. 
We show participants' post-trial questionnaire responses on trust in robotics and satisfaction to the robot in \autoref{fig:trust & stat}. 

Participants rated the animated displays ($avg=7.9$) significantly more trustworthy than static displays ($avg=6.1$), confirmed by a Welch’s t-test $t = 4.7631 $, $ p < 0.001 $. 
Participants were significantly more satisfied with animated displays ($avg= 8.347$) than static displays ($avg = 6.857$), validated by a Welch’s t-test $t =4.026 $, $ p < 0.001$. 
From the results of our study we found that static eyes provided the best cues for task completion success. We also found that the robot with animated expressions engendered  significantly higher trust and satisfaction compared to the robot with static expressions. Our results also showed that participants interpreted the static eyes the best, while their  self-reported understanding of the displays had a low correlation with their actual Interpretation Game performance.



\section{Discussion}
We undertook this study to examine whether animation and anthropomorphism, which are at the forefront of robot technology, or the use of standard and familiar static icons provide the best means of robot-to-human communication. The results of our study found that our participants had the highest levels of trust and satisfaction when they worked with a robot with animated displays. In contrast, participants who worked with a robot with basic static eyes had the highest pass-to-fail rate and task completion success. In addition, our participants were able to most accurately interpret the static icons.  We also found that participants’ self-reported understandability of the displays was not highly correlated with their objective understanding of the displays.

Altogether, the results suggest that different visual cues have different advantages and that it is important to measure the communication signals in both objective and subjective manners. Because trust, understanding, clarity and efficiency are all necessary components in HRI, we believe that no one modality should take precedence to the exclusion of the others. Rather, a combination of animation and static icons or anthropomorphism and static icons may be ideal to bridge the gap in robot-to-human communication.

One limitation of this study is that we compared the modalities using only one robot and a single task. We believe, however, that our conclusions are generally applicable to other robots equipped with LED screens, as screen-based displays are widely adopted in both commercial and research-oriented robots, particularly those designed for human interaction. Including another robot, especially a humanoid, could further improve the validity of our findings. The study could also benefit from incorporating an additional task, such as a manipulation task, to broaden the insight of our work. 

\section{Conclusion}
In this work, we conducted a public space study with non-expert participants to explore how four non-verbal display modalities affect human-robot communication in a collaborative navigation task.  Our purpose was to investigate whether animation and anthropomorphism should be favored over basic static icons long-used in robot design.  Our results showed, however, that participants with the robot equipped with static eyes achieved the highest pass-to-fail ratio, and participants interpreted static icons the best. 
Nevertheless, participants rated the robot with animated displays as more trustworthy and satisfactory.
Moreover, we found that their self-reported understandability actually had a low correlation with their objective understandability, suggesting the importance of including both subjective and objective  measurements.
Our findings have the potential to help guide robot
design towards robots that are not only useful, but also intuitively
understood by people across all backgrounds. 

\newpage

\bibliographystyle{plain}
\bibliography{bibtex}

\vspace{12pt}

\end{document}